# A Method for Self-Service Rehabilitation Training of Human Lower Limbs*

Zhaowen Shao[1], Jun Li[1], *Senior Member*, *IEEE*, and Lingtao Yu[2]

*Abstract*—In recent years, the research of rehabilitation robot technology has become a hotspot in the field of rehabilitation medicine engineering and robotics. To assist active rehabilitation in patients with unilateral lower extremity injury, we propose a new self-service rehabilitation training method to control the injured lower extremity through its contralateral healthy upper limbs. Firstly, the movement data of upper limbs and lower limbs of healthy people in normal walking state are obtained by gait measurement experiment. Secondly, the eigenvectors of upper limb and lower limb movements in a single movement cycle are extracted respectively. Thirdly, the linear mapping relationship between the upper limbs movement and the lower limbs movement is identified using the least squares method. Finally, the simulation experiment of self-service rehabilitation training is implemented on MATLAB/Simulink. The results indicate that the identified linear mapping model can achieve good accuracy and adaptability. The self-service rehabilitation training method is effective for helping patients with unilateral limb injury to make rehabilitation training on themselves.

## I. INTRODUCTION

Recent results in rehabilitation medicine indicated that the brain has plasticity and neural network remodeling mechanisms [1-3]. The preconditions of this mechanism were: 1) sufficient motion and sensory stimulation; 2) active response. The clinical application of rehabilitation training robot manifested that the higher the initiative and enthusiasm of patients in the process of rehabilitation training, the better the effect of rehabilitation training [4-6].

The existing rehabilitation training methods for paralysed patients were mainly divided into passive training and active training. Lv et al. [7] exploited computed torque method to study the passive rehabilitation training of paralysed patients. However, the inadequacy of patients' initiative during passive rehabilitation training affected the effect of training. The existing active training mainly realized by mechanical signals and biomedical signals. For example, hybrid position/force control [8-9] and impedance control [10-13] were exploited to study human-machine interaction, and sEMG signals was applied to estimate muscle strength [14-16] or EEG was applied to identify human motion intentions [17-19]. Yu et al. [20] exploited the fuzzification method to establish the master-slave mapping relationship between upper limb movement and lower limb movement. The upper limb movement data was applied as input to control the lower limb for rehabilitation training.

There are still some problems in the existing rehabilitation training methods, which are not conducive to the active involvement of patients in rehabilitation training. Rehabilitation training based on mechanical signals requires high accuracy of dynamic modeling. However, the interaction between human and machine makes the rehabilitation robot time-varying uncertainties, which makes it impossible to accurately establish the dynamic model of the human-machine system, and makes the traditional model-based control method difficult to achieve effective control. The sEMG signals and EEG used in biomedical signal-based rehabilitation training have problems of low signal-to-noise ratio and signal instability. The master-slave mapping relationship between upper limb and lower limb movement established by Yu [20] on the basis of curve envelope is not good enough for upper limb movement in different amplitude ranges.

In view of this situation, considering the correlation between upper limbs and lower limbs movements during normal walking, this paper studies a new active rehabilitation training method: self-service rehabilitation training method. This method controls the injured lower limb through its contralateral healthy upper limb.

In this paper, the movement data of upper and lower limbs of healthy people in normal walking state are obtained by gait measurement experiment. The mapping relationship between upper limb movement and lower limb movement was established. On the basis of the mapping relationship, the upper limb is used to control the lower limbs for rehabilitation training. The work of this paper is expected to help patients with unilateral limb injuries to achieve self-control of the rehabilitation training process, improve the adaptability of the rehabilitation system for upper limb movements in different amplitude ranges, and then improve the initiative and enthusiasm of patients for rehabilitation training.

## II. REHABILITATION TRAINING PROCESS

During self-help rehabilitation training, patients are placed on weight-reducing brackets to ensure their safety, reduce part of the patients' weight and reduce the training burden. The upper limb of the healthy side of the patient can move freely, and the lower limbs are connected with the corresponding exoskeleton rehabilitation equipment. By measuring the movement data of upper limbs through sensors, the movement data of upper limbs are mapped to the movement data of the lower limbs through certain methods, which can be used to control the movement of lower limbs exoskeletons, and then drives the lower limbs for rehabilitation training.

*This work was supported in part by the National Natural Science Foundation of China under Grant 61773115, the Jiangsu Province Natural Science Foundation under Grant BK20161427.

[1]Z. Shao and J. Li are with the Ministry of Education Key Laboratory of Measurement and Control of CSE, Southeast University, Nanjing 210096, China (e-mail: 230179719@seu.edu.cn; j.li@seu.edu.cn).

[2]L. Yu is with the College of Mechanical and Electrical Engineering, Harbin Engineering University, Harbin 150001, China (e-mail: yulingtao@163.com).

For the convenience of analysis, the gait motion is decomposed into several separate periods for processing. Eigenvectors are further extracted on the basis of the single cycle data to represent the corresponding limb movements in the current cycle. The implementation plan of self-service rehabilitation training is displayed in Fig. 1. The scheme first extracts the eigenvectors of the upper limb movements data in single cycles, then acquires the lower limb movements eigenvectors through mapping relationship, and then obtains the lower limb motion data in single cycle through feature restoration processing, which is then applied to control the lower limb movements. The program for establishing the mapping relationship between upper limb and lower limb movements is displayed in Fig. 2.

The gait measurement experiment is exploited to gain the motion data of the upper limbs and lower limbs of the human body under normal walking conditions. Then the eigenvectors are abstracted from the movement data of upper limbs and lower limbs in separate cycles. The mapping relationship between upper limb and lower limb movements is established by analyzing the eigenvectors of upper limb and lower limb movements in separate cycles.

In the process of self-service rehabilitation training, the ankle joint is a passive joint, and its movement state is determined by the lower limbs and the ground. The state of movement of the lower limbs lags behind that of the upper limbs by one gait cycle.

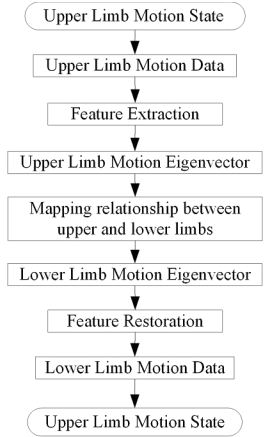

Fig. 1. Implementation scheme of self-service rehabilitation training

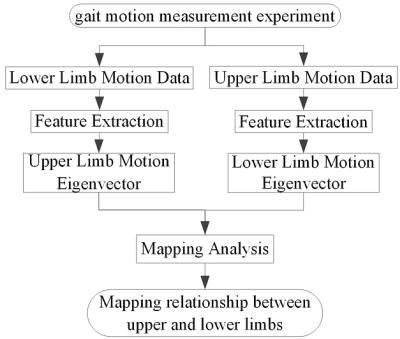

Fig. 2. Scheme for establishing the mapping relationship

## III. GAIT MOVEMENT DATA ACQUISITION AND DATA ANALYSIS

### A. Gait Movement Measurement Experiment

The FAB real-time wireless motion capture and mechanical evaluation system is exploited to gather experimental data of single person and multiple groups, and the sampling frequency is 100 Hz. It is assumed that the motion phases of the upper limbs and the lower limbs of both sides are different by half a cycle when the human body is walking normally. The rotation data of right upper limb shoulder and elbow joints are applied as upper limb movement data, and the rotation data of left lower limb hip and knee joints are applied as lower limb movement data.

### B. Upper Limb and Lower Limb Movement Feature Extraction

Through analysis, it is found that the peak and trough of each joint angle curve of upper and lower limbs in a single cycle are relatively representative for the corresponding curve. As long as the peak and trough are determined, the corresponding curve can be confirmed to a large extent. Therefore, the peak and trough amplitudes of the joint rotation curves of upper and lower limbs in separate cycles are selected as motion features, and then integrated into corresponding eigenvectors.

During normal walking, there are some disturbances in the motion data of upper and lower limbs, which may interfere with the feature extraction of motion data. The disturbance of motion data can be characterized as a sudden and large change in the rate of curve change. Based on statistical regularity, band-pass filters for the rate of variation of joint angle curves of upper and lower limbs are constructed respectively, as shown in Fig. 3. In the process of extracting motion features, the data whose change rate is not in the range of band-pass filters are regarded as disturbances.

In order to abstract the motion characteristics of separate cycles, the time intervals of peak and trough appearing in the motion curves of upper and lower limbs of single periods are firstly computed. In this time interval, the peaks and troughs that match the band-pass filter are selected and integrated into corresponding feature vectors.

The eigenvectors of the upper limb movements in separate cycles are denoted as $x_i = (x_{i1}, x_{i2}, x_{i3}, x_{i4})^T$, and its four components are the first trough and peak of the shoulder joint curves and the first trough and peak of the elbow joint curves. The eigenvectors of lower limb movements in separate cycles are denoted as $y_i = (y_{i1}, y_{i2}, y_{i3}, y_{i4})^T$, and its four components are the first trough and peak of the hip joint curves and the first peak and trough of the knee joint curves. The set of upper limb movements eigenvectors of multiple cycles is $X = \{x_i\}$. The set of lower limb movements eigenvectors of several corresponding cycles is $Y = \{y_i\}$.

The distribution of the eigenvectors set of the single-cycle upper and lower limb movements is displayed in Fig. 4. For ease of display, the eigenvectors in Figure 4 are divided into 4 clusters, respectively.

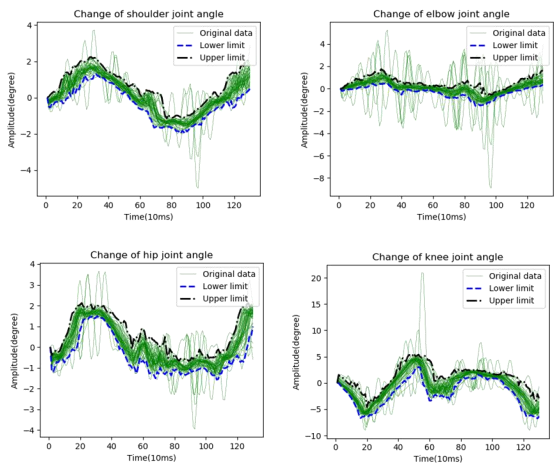

Fig. 3. Change rate of joint angle curve and corresponding bandpass filter

## IV. MODELING OF LOWER LIMB MOVEMENT BASED ON UPPER LIMB MOVEMENT

### A. Model Class Selection

It is assumed that the mapping relationship between upper and lower limb movements is a function $f$. Then the mapping relationship between $X=\{x_i\}$ and $Y=\{y_i\}$ can be expressed as the form of piecewise functions as follows:

$$y=\{f_i(x),\ x\in B(x_i,\varepsilon_i)\},\ x\in X, y\in Y. \quad (1)$$

Among them, $\varepsilon_1,\varepsilon_2,\cdots,\varepsilon_m$ are the corresponding neighborhood parameters and $m$ is the number of samples in the set of eigenvectors.

Similar upper limb input should correspond to similar lower limb output in order to facilitate self-service rehabilitation training for patients. For linear systems, similar inputs produce similar outputs. For ease of application, the piecewise functions is simplified to the low-dimensional linear mapping, namely:

$$T:X\to Y, y_i=Tx_i+b. \quad (2)$$

Where $T$ is a low-dimensional linear operator, $x_i\in R^{4\times 1}$, $y_i\in R^{4\times 1}$, $b\in R^{4\times 1}$.

### B. Identification of Model Parameters Using the Least Squares Method

*1) Identification of linear mapping model based on least squares method*

For the set of eigenvectors of upper limb and lower limb movements, the linear model of their mapping relationship is identified by least squares method.

From mapping relation, we can get:

$$(y_1,y_2,...,y_k)=T(x_1,x_2,...,x_k)+(b,b,...,b). \quad (3)$$

Let $T=\begin{pmatrix} t_{11} & t_{12} & t_{13} & t_{14} \\ t_{21} & t_{22} & t_{23} & t_{24} \\ t_{31} & t_{32} & t_{33} & t_{34} \\ t_{41} & t_{42} & t_{43} & t_{44} \end{pmatrix} \overset{\Delta}{=} \begin{pmatrix} t_1 \\ t_2 \\ t_3 \\ t_4 \end{pmatrix}$,

thus, $y_{ij}=x_i^T\cdot t_j^T+b_j$, where, $i=1,2,\cdots,m$; $j=1,2,3,4$.

Remember $Y_j=\begin{pmatrix} y_{1j} \\ \vdots \\ y_{ij} \\ \vdots \\ y_{mj} \end{pmatrix}$, $\Phi=\begin{pmatrix} x_1^T & 1 \\ \vdots & \vdots \\ x_i^T & 1 \\ \vdots & \vdots \\ x_m^T & 1 \end{pmatrix}$,

thus $Y_j=\Phi\cdot t_j^T$.

The estimated value of $(t_j^T,b_j)$ based on the least squares method is:

$$(\hat{t}_j^T,\hat{b}_j)=(\Phi^T\Phi)^{-1}\cdot\Phi^T Y^j. \quad (4)$$

Thus, the least squares estimate of the linear operator $T$ is:

$$\hat{T}=\begin{pmatrix} \hat{t}_1 \\ \hat{t}_2 \\ \hat{t}_3 \\ \hat{t}_4 \end{pmatrix}=\begin{pmatrix} 0.0946 & -0.1917 & 1.6623 & -0.0483 \\ -0.0151 & -0.1066 & 1.3587 & 0.0959 \\ 0.2514 & -0.1003 & -1.5821 & 0.2368 \\ 0.0907 & -0.1945 & -1.0221 & -0.0946 \end{pmatrix}. \quad (5)$$

The least squares estimate of the initial value $b$ is:

$$\hat{b}=\begin{pmatrix} \hat{b}_1 \\ \hat{b}_2 \\ \hat{b}_3 \\ \hat{b}_4 \end{pmatrix}=\begin{pmatrix} 3.0487 \\ 40.2068 \\ -3.3855 \\ -89.9575 \end{pmatrix}. \quad (6)$$

*2) Error analysis of linear mapping model*

In this paper, the mean and standard deviation of the residual between the mapping eigenvector $y_i'$ and the original eigenvector $y_i$ are exploited to reflect the proximity of them, and then the error of the output data of the linear mapping model relative to the original data is reflected. The mean and standard deviation of the residuals of the components of the lower limb movement eigenvector are displayed in Table Ⅰ.

The mean values of the residuals of each component of the lower limb movement eigenvectors is nearly zero, which indicates that the estimated value is unbiased. The residuals of these eigenvector components reflect the non-linear factors in the movement correlation between the upper limbs and lower limbs. Overall, these errors are acceptable.

## V. SIMULATION EXPERIMENT AND RESULT ANALYSIS OF SELF-SERVICE REHABILITATION TRAINING

According to the existing conditions, the self-service rehabilitation training process is simulated through simulation experiments, and then the mapping relationship between upper and lower limb movements established before is confirmed. The model effect of the self-service rehabilitation training simulation experiment by MATLAB/Simulink is displayed in

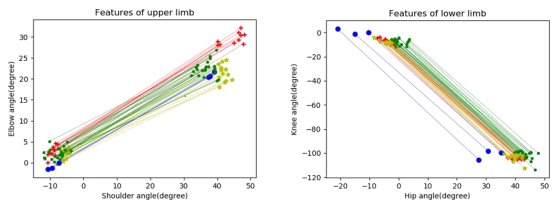

Fig. 4. Distribution of eigenvectors extracted from joint angle curves

TABLE I. ERROR OF LINEAR MAPPING MODEL

| Model error | y₋1 | y₋2 | y₋3 | y₋4 |
|---|---|---|---|---|
| Mean(degree) | -2.6e-4 | -2.7e-5 | -4.2e-4 | -1.5e-3 |
| Std(degree) | 3.4532 | 2.9461 | 2.0199 | 2.6823 |

Fig. 5.

### A. Feature Extraction and Feature Mapping

In the simulation environment, the valleys and peaks of shoulder and elbow rotation curves are abstracted as feature points. Considering the randomness of upper limb movements and the possible jitter during walking, a set of judgment conditions are set in the process of feature extraction. Only when the points on the joint rotation curves satisfy these judgment conditions at the same time can they be determined as corresponding feature points.

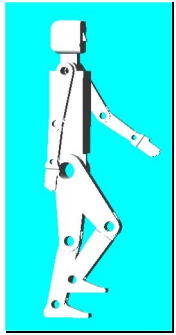

Fig. 5. Model effect of simulation experiment

The amplitudes of each feature point in a single cycle are integrated into the eigenvector $x_i$ of the upper limb motion by a certain method. In the process of self-service rehabilitation training, upper limb movement is the input and lower limb movement is the output. For any upper limb motion eigenvector $x_i$, it is mapped to the corresponding lower limb motion eigenvector $y'_i$ to control the lower limb movement state. Namely:

$$y'_i = Tx_i + b . \quad (7)$$

### B. Feature Restoration

Feature restoration is applied to transform the lower limb motion eigenvector $y'_i$ into the actual rotation angles of the lower limb joints. Therefore, it is necessary to establish the corresponding relationship between the eigenvector $y'_i$ and the respective joint angle curves of the lower limb movement.

This paper implements feature restoration by reference vectors and reference curves. In order to acquire a group of representative reference vectors and reference curves, it is necessary to cluster the feature vectors of upper and lower limb movements. Set $X$ and set $Y$ are merged and divided into nine clusters by KMeans clustering algorithm. Four representative clusters are chosen from them. The average vectors of each cluster are taken as reference vectors, and the average curves of each cluster are taken as reference curves.

The mean vectors $\bar{y}_1, \bar{y}_2, \bar{y}_3, \bar{y}_4$ of the four clusters is as follows:

$$\begin{cases} \bar{y}_1 = (-4.0344 \quad 39.8672 \quad -8.0813 \quad -102.6813)^T \\ \bar{y}_2 = (-0.9446 \quad 43.0250 \quad -5.5536 \quad -99.4214)^T \\ \bar{y}_3 = (-3.3208 \quad 42.1188 \quad -7.9562 \quad -105.1646)^T \\ \bar{y}_4 = (3.1625 \quad 45.8825 \quad -8.2300 \quad -102.1600)^T \end{cases} . \quad (8)$$

Calculate the average curve of each cluster: $L_{\bar{y}_1}, L_{\bar{y}_2}, L_{\bar{y}_3}, L_{\bar{y}_4}$, and fit them into the corresponding functional relationships: $f_{\bar{y}_1}(t), f_{\bar{y}_2}(t), f_{\bar{y}_3}(t), f_{\bar{y}_4}(t)$.

Suppose that the lower limb motion eigenvector $y'_i$ acquired from the mapping satisfies the following relationship:

$$y'_i = a_1\bar{y}_1 + a_2\bar{y}_2 + a_3\bar{y}_3 + a_4\bar{y}_4 . \quad (9)$$

Among them, $a_1, a_2, a_3, a_4$ represent the corresponding weight coefficients.

Solve the equation: $y'_i = a_1\bar{y}_1 + a_2\bar{y}_2 + a_3\bar{y}_3 + a_4\bar{y}_4$, and get the coefficients $a_1, a_2, a_3, a_4$.

Thus, the function relation of the lower extremity joint rotation curve $L'_{y_i}$ output by the system is as follows:

$$f'_{y_i} = a_1 f_{\bar{y}_1}(t) + a_2 f_{\bar{y}_2}(t) + a_3 f_{\bar{y}_3}(t) + a_4 f_{\bar{y}_4}(t) . \quad (10)$$

The gained angular curves of each joint of lower limb motion are applied to control the motion state of lower limbs. The state of movement of lower limbs lags behind that of upper limbs by one gait cycle.

### C. Analysis of Simulation Experiment Results

In the process of self-service rehabilitation training, the lower limb motion state of the system output needs to meet the following two requirements: (1) The lower limb motion state of the system output can largely restore the original lower limb motion state. (2) The lower extremity motion state of the system output keeps in harmony with the upper extremity motion state. Therefore, the results of simulation experiments need to be analyzed from two aspects.

#### 1) Error analysis of lower limb movement restoration

The error of lower limb motion restoration is reflected by the deviation between the lower limb motion state output by the system and the original lower limb motion state of the previous cycle. It is mainly divided into phase error and amplitude error. The lower limb motion data output by the system is translated forward by one cycle and compared with the original lower limb motion data as shown in Fig. 6.

The phase error in Fig. 6 is due to the randomness of human walking posture. The length of each gait cycle is different when the human body walks. The time period of the lower limb motion curve output during the feature reduction is fixed. As a result, there will always be a certain degree of phase deviation between the output curve and the original curve. The phase error between the output data and the original data of lower limb motion will be reduced gradually when the motion state of upper limb in each cycle tends to be stable. The errors of lower limb motion restoration during the three experimental procedures are displayed in Table Ⅱ. Generally speaking, the deviations between the lower extremity motion curve output by the system and the original curve are mainly due to the phase error.

*2) Coordination analysis of upper limb and lower limb movements*

The coordination of upper and lower limb movements is reflected by the phase difference between them in the current cycle. The corresponding relationship between the lower extremity motion curves output by the system and the actual upper extremity motion curves during the simulation experiments are displayed in Fig. 7.

The phase differences between the upper limb and the lower limb movement states of the upper limbs and the lower limbs are relatively coordinated.

## VI. CONCLUSION

This paper researches a self-service rehabilitation training method to control the injured lower limbs of the opposite side through one side healthy upper limbs. The results indicate that the linear mapping model identified by least squares method has good accuracy and adaptability to upper limb movements in different amplitude ranges. The lower limb motion state output by the system is close to its original state and can keep in harmony with the upper limb motion state. The self-service rehabilitation training method researched in this paper can better restore the lower limbs movement state through the upper limb movement state, and then help the patients with unilateral limbs injury to realize the self-control of the rehabilitation training process.

The method in this paper is to establish a mapping relationship between the amplitude of upper limb movement and lower limb movement. Therefore, only the amplitude

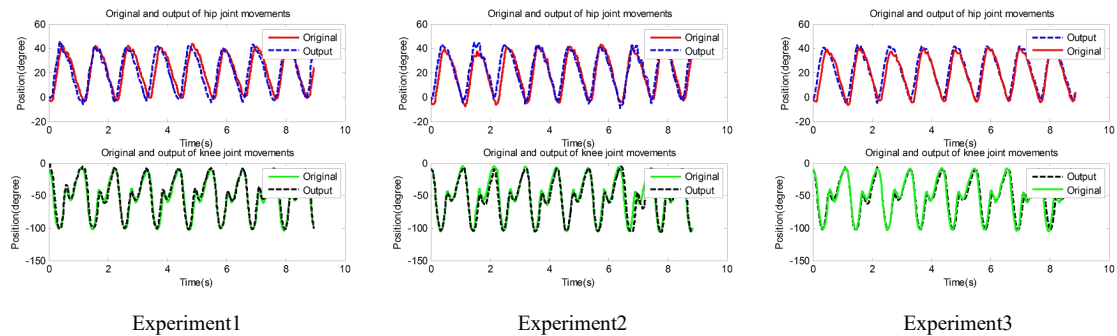

Experiment1　　　　　　　　　　　　Experiment2　　　　　　　　　　　　Experiment3

Fig. 6. Comparison of the output data with the original data of lower limb movements

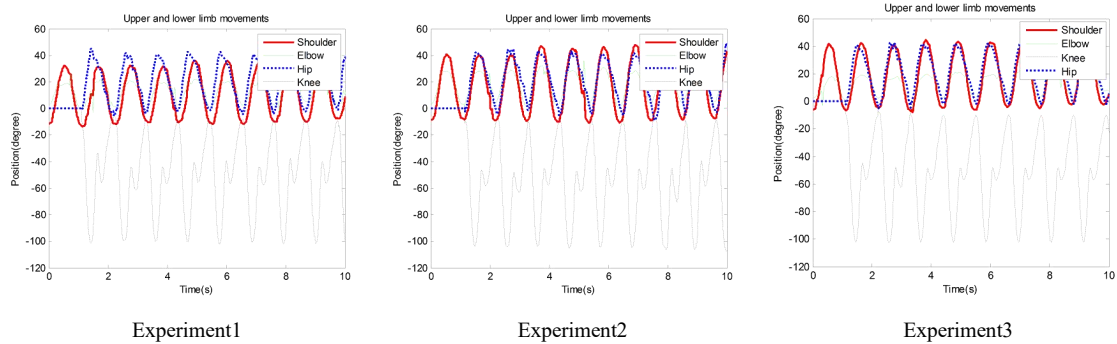

Experiment1　　　　　　　　　　　　Experiment2　　　　　　　　　　　　Experiment3

Fig. 7. Motion curves of lower limbs and upper limbs during simulations

limb movements in the raw data was calculated to gain its distribution in the normal walking state. The phase difference between the lower extremity movement output by the system during the simulation experiment and the upper extremity movement in the current cycle is calculated, as displayed in Table Ⅲ.

Overall, the distribution of the phase difference between the upper limb and the lower limb during the simulation experiment is close to that in the original data. The animation during the simulation experiment reveals that the movement states of the lower limbs are relatively natural, and the relationship between upper limb movement and lower limb movement can be guaranteed to match as much as possible. For the case of large phase difference between upper limb movements in different periods, it will lead to the uncoordinated phase between upper limb and lower limb movements, and then lead to the uncoordinated training posture. Therefore, the self-service rehabilitation training method studied in this paper requires that the upper limbs on the healthy side of patients have certain independent motor ability.

TABLE II. ERROR OF LOWER LIMB MOTION RESTORATION

| Data | Error | | Mean(degree) | Std(degree) |
|---|---|---|---|---|
| Experiment1 | Phase | | -0.0378 | 0.0383 |
| | Amplitude | Hip | 0.1888 | 7.1506 |
| | | Knee | -2.1223 | 6.1905 |
| Experiment2 | Phase | | -0.0300 | 0.0141 |
| | Amplitude | Hip | -2.1999 | 6.1676 |
| | | Knee | 2.8640 | 8.8241 |
| Experiment3 | Phase | | -0.0456 | 0.0113 |
| | Amplitude | Hip | -2.6807 | 4.7426 |
| | | Knee | 1.1271 | 8.1681 |

TABLE III. PHASE DIFFERENCE BETWEEN UPPER AND LOWER LIMB MOVEMENT

| Phase difference | Original | Experiment 1 | Experiment 2 | Experiment 3 |
|---|---|---|---|---|
| Mean(degree) | 0.0254 | -0.0611 | 0.0433 | -0.0089 |
| Std(degree) | 0.0242 | 0.0348 | 0.0250 | 0.0093 |